\documentclass[11pt,a4paper]{article}
\usepackage[hyperref]{acl2020}
\usepackage{times}
\usepackage{latexsym}

\usepackage{microtype}
\usepackage{url}
\usepackage{import}
\usepackage{multirow}
\usepackage{multicol}
\usepackage{makecell}
\usepackage{booktabs}
\usepackage{xcolor}
\usepackage{colortbl}
\usepackage{subcaption}
\usepackage{amsmath}

\usepackage{graphicx}
\usepackage{enumitem}
\aclfinalcopy 
\def\aclpaperid{2257} 
\newcommand{\minorchange}[1]{#1}

\title{iSarcasm: A Dataset of Intended Sarcasm}

\author{Silviu Vlad Oprea \\
  School of Informatics\\
  University of Edinburgh\\
  Edinburgh, United Kingdom\\
  {\tt silviu.oprea@ed.ac.uk} \\\And
  Walid Magdy \\
  School of Informatics\\
  University of Edinburgh \\
  Edinburgh, United Kingdom\\
  {\tt wmagdy@inf.ed.ac.uk} \\}

\date{}
\newcommand{\q}[1]{\bgroup\color{red}[QUESTION: #1]\egroup}
\newcommand{\walid}[1]{\bgroup\color{blue}[Walid: #1]\egroup}
\newcommand{\bonnie}[1]{\bgroup\color{blue}[Bonnie: #1]\egroup}
\newcommand{\todo}[1]{\bgroup\color{blue}[TODO: #1]\egroup}
\newcommand{\towalid}[1]{\bgroup\color{red}[@WALID: #1]\egroup}
\newcommand{\place}{\bgroup\color{green!70!black}[PLACEHOLDER]\ \egroup}

\begin{document}
\maketitle
\begin{abstract}
We consider the distinction between intended and perceived sarcasm in the context of textual sarcasm detection. The former occurs when an utterance is sarcastic from the perspective of its author, while the latter occurs when the utterance is interpreted as sarcastic by the audience. We show the limitations of previous labelling methods in capturing intended sarcasm and introduce the iSarcasm dataset of tweets labeled for sarcasm directly by their authors. 
Examining the state-of-the-art sarcasm detection models on our dataset showed low performance compared to previously studied datasets, which indicates that these datasets might be biased or obvious and sarcasm could be a phenomenon under-studied computationally thus far. By providing the iSarcasm dataset, we aim to encourage future NLP research to develop methods for detecting sarcasm in text as intended by the authors of the text, not as labeled under assumptions that we demonstrate to be sub-optimal.

\end{abstract}
%
%
\section{Introduction\footnote{\textbf{This article is a preprint of an article accepted for publication at ACL 2020.}}}
Sarcasm is a form of irony that occurs when there is some discrepancy between the literal and intended meanings of an utterance.
This discrepancy is used to express dissociation towards a previous proposition, often in the form of contempt or derogation~\citep{WILSON20061722}.
Sarcasm is omnipresent in social media text and can be highly disruptive of systems that harness this data for sentiment and emotion analysis~\citep{L14-1527}.
It is therefore imperative to devise models for sarcasm detection.
The effectiveness of such models depends on the availability and quality of labelled data used for training.
Collecting such data is challenging due to the subjective nature of sarcasm. For instance, \citet{Context3} notice a lack of consistence in how sarcasm is used by people of different socio-cultural backgrounds. As a result, an utterance \emph{intended} sarcastic by its author might not be \emph{perceived} as such by audiences of different backgrounds~\citep{Context2,oprea-2020-cscw}.

There are two methods used so far to label texts for sarcasm: distant supervision, where texts are considered sarcastic if they meet predefined criteria, such as including specific hashtags; and manual labelling by human annotators.
We believe both methods are sub-optimal for capturing the sarcastic intention of the authors of the texts. As a result, existing models trained on such datasets might be optimized to capture the noise induced by these labelling methods.

In this paper, we present the iSarcasm dataset of tweets labelled for sarcasm by their authors. To our knowledge, this is the first attempt to create noise-free examples of intended sarcasm. In a survey, we asked Twitter users to provide both sarcastic and non-sarcastic tweets that they had posted in the past.
For each sarcastic tweet, we asked them to explain why it was sarcastic and how they would convey the same meaning non-sarcastically. Labels were thus implicitly specified by the authors themselves. 
We implemented restrictive quality control to exclude spurious survey responses. We then asked a trained linguist to manually check the sarcastic tweets and further label them into the subcategories of sarcasm defined by~\citet{sarcasm-categories}.

We further collected third-party sarcasm labels for the tweets in iSarcasm from workers on a crowdsourcing platform. Third-party annotation for sarcasm has been conducted before~\cite{filatova,Riloff,abercrombie2}, but no studies checked the ability of the annotators to capture the actual sarcasm meant by the authors. On iSarcasm, annotators recognise author labels with an F-score of 0.616. This indicates that sarcasm is a subjective phenomenon, challenging even for humans to detect. Further, it demonstrates that using third-party annotators to label texts for sarcasm can lead to inaccurate labels.

We implemented state-of-the-art sarcasm detection models~\cite{tay-att,cascade,van-hee-etal-2018-semeval} and tested them on iSarcasm, to investigate their effectiveness in capturing sarcasm as intended by the authors. While these models achieve F-scores reaching 0.874 on existing datasets, they yield a maximum F-score of 0.364 on iSarcasm, suggesting that previous datasets might be biased or obvious.
This highlights the importance of developing new approaches for sarcasm detection that are more effective at capturing author intention.

iSarcasm contains 4,484 English tweets, each with an associated intended sarcasm label provided by its author, with a ratio of roughly 1:5 of sarcastic to non-sarcastic tweets. Each sarcastic tweet has an extra label indicating the category of sarcasm it belongs to. We publish the dataset publicly for research purposes\footnote{https://github.com/silviu-oprea/iSarcasm}.
%
%
\section{Background}
\subsection{Intended and Perceived Sarcasm}
\label{section:background:linguistics}
%
The way sarcasm is used can vary across socio-cultural backgrounds. \citet{Context3} notice that members of collectivist cultures tend to express sarcasm in a more subtle way than individualists. They also point out gender differences. Females seem to have a more self-deprecating attitude when using sarcasm than males. \citet{Context2} find some cultures to associate sarcasm with humour more than others. There are also cultures who do not use sarcasm at all, such as the Hua, a group of New Guinea Highlanders~\citep{talkischeap}.
Because of these differences, an utterance intended sarcastic by its author might not be perceived as such by the audience~\citep{Jorgensen1984}. Conversely, the audience could perceive the utterance as sarcastic, even if it was not intended as such. 

The distinction between intended and perceived sarcasm, also referred to as encoded and decoded sarcasm, respectively, has been pointed out in previous research~\citep{kaufer,Context2}. However, it has not been considered in a computational context thus far when building datasets for textual sarcasm detection.
We believe accounting for it is essential, especially nowadays. Consider social media posts that can reach audiences of unprecedented sizes. It is important to consider both the communicative intention of the author, for tasks such as opinion mining, as well as possible interpretations by audiences of different sociocultural backgrounds, for tasks such as hate-speech detection.
\subsection{Sarcasm Datasets}
\label{section:background:datasets}
Two methods were used so far to label texts for sarcasm: distant supervision and manual labelling.
\paragraph{Distant supervision}
This is by far the most common method. Texts are considered positive examples (sarcastic) if they meet predefined criteria, such as containing specific tags, such as \#sarcasm for Twitter data~\cite{ptacek}, and /s for Reddit data~\cite{khodak-2017}, or being posted by specific social media accounts~\citep{barbieri2014italian}. Negative examples are usually random posts that do not match the criteria.
Table~\ref{table:datasets} gives an overview of datasets constructed this way, along with tags or accounts they associate with sarcasm.

The main advantage of distant supervision is that it allows building large labelled datasets with no manual effort.
However, as we discuss in Section~\ref{section:motivation}, the labels produced can be very noisy.
%
\renewcommand{\arraystretch}{1.1}
\begin{table*}[t]
    \centering
    \small
    \begin{tabular}{@{}p{6.5cm}cp{6cm}@{}}
        \toprule
        \textbf{Sarcasm labeling method}  & \textbf{Source} & \textbf{Details / Tags / Accounts}  \\
        \midrule
        \textbf{Distant supervision}          &                 &                   \\
            \citet{davidov2010}  & Twitter         & \#sarcasm, \#sarcastic, \#not \\
            \citet{barbieri20141} & Twitter        & \#sarcasm, \#education, \#humor, \#irony, \#politics\\
            \citet{ptacek}        & Twitter        & \#sarcasm, \#sarcastic, \#irony, \#satire\\
            \citet{firstcontextualised,joshi-2015} & Twitter        & \#sarcasm, \#sarcastic\\
            \citet{gonzalez,reyes2012,liebrecht2013,bouazizi-2015,bharti-2015}     &     Twitter     & \#sarcasm\\
            \citet{barbieri2014italian} & Twitter  & tweets posted by \emph{@spinozait} or \emph{@LiveSpinoza} \\
            \citet{khodak-2017}   & Reddit         & /s\\
        \midrule
        \textbf{Manual annotation / Hybrid}            &                 &                   \\
            \citet{Riloff,benamara2017analyse,cignarella2018overview,van-hee-etal-2018-semeval,Bueno2019OverviewOT}       & Twitter         & tweets            \\
            \citet{abercrombie2} & Twitter         & tweet-reply pairs \\
            \citet{filatova}     & Amazon          & product reviews   \\
        \bottomrule
    \end{tabular}
    \caption{Datasets previously suggested for sarcasm detection, all annotated using either distant supervision or manual labelling, as discussed in Section~\ref{section:background:datasets}.}
    \label{table:datasets}
\end{table*}
\paragraph{Manual labelling}
An alternative to distant supervision is collecting texts and presenting them to human annotators for labelling.
\citet{filatova} asks annotators to find pairs of Amazon reviews where one is sarcastic and the other one is not, collecting 486 positive and 844 negative examples.
\citet{abercrombie2} annotate 2,240 Twitter conversations, ending up with 448 positive and 1,732 negative labels, respectively.
\citet{Riloff} use a hybrid approach, where they collect a set of 1,600 tweets that contain \#sarcasm or \#sarcastic, and another 1,600 without these tags. They remove such tags from all tweets and present the tweets to a group of human annotators for final labelling. We call this the \emph{Riloff dataset}.
A similar approach is employed by~\citet{van-hee-etal-2018-semeval} who recently presented their dataset as part of a SemEval shared task for sarcasm detection. It is a balanced dataset of 4,792 tweets. We call it the \emph{SemEval-2018 dataset}.
\subsection{Sarcasm Detection Models}
\label{section:background:models}
Based on the information considered when classifying a text as sarcastic or non-sarcastic, we identify two classes of models across literature: text-based models and contextual models.
\paragraph{Text-based models}
These models only consider information available within the text being classified. Most work in this direction considers linguistic incongruity~\citep{campbell2012there} to be a marker of sarcasm. \citet{Riloff} look for a positive verb in a negative sentiment context. \citet{r2} search for a negative phrase in a positive sentence. \citep{farias2015valento} measure semantic relatedness between words using WordNet-based similarity. \citet{riloffembeddings} use the cosine similarity between word embeddings. Recent work~\citep{tay-att} uses a neural intra-attention mechanism to capture incongruity.
\paragraph{Contextual models}
These models utilize information from both the text and the context of its disclosure, such as author information. There is a limited amount of work in this direction.
Using Twitter data, \citet{firstcontextualised} represent author context as manually-curated features extracted from their historical tweets.
\citet{amir} merge all historical tweets into one document and use the Paragraph Vector model~\citep{doc2vec} to build an embedding of that document.
Building on this, \citet{cascade} extract additional personality features from the merged historical tweets with a model pre-trained on a personality detection corpus. Using the same strategy, \citet{oprea-2019} build separate embeddings for each historical tweet and identify author context with their the weighted average.

Despite reporting encouraging results, all previous models are trained and tested on datasets annotated via manual labelling, distant supervision, or a mix between them. We believe both labelling methods are limited in their ability to capture sarcasm in texts as intended by the authors of the texts without noise. We now discuss how noise can occur.
%
%
\section{Limitations of Current Labelling Methods}
\label{section:motivation}
In this section, we discuss limitations of current labelling methods that make them sub-optimal for capturing intended sarcasm.
We demonstrate them empirically on the Riloff dataset~\cite{Riloff}, which uses a hybrid approach for labelling.
\subsection{Limitations of Distant Supervision}
\label{section:motivation:distant-supervision}
Since it is based on signals provided by the authors, distant supervision might seem like a candidate for capturing intended sarcasm. However, we identify a few fundamental limitations with it.
First, the tags may not mark sarcasm, but may constitute the subject or object of conversation, e.g. \emph{\#sarcasm annoys me!}. This could lead to false positives.
Second, when using tags such as \#politics and \#education~\cite{barbieri20141}, there is a strong underlying assumption that these tags are accompanied by sarcasm, potentially generating further false positives.
The assumption that some accounts always generate sarcasm~\cite{barbieri2014italian} is similarly problematic.
In addition, the intended sarcasm that distant supervision does capture might be of a specific flavor, such that, for instance, the inclusion of a tag would be essential to ensure inferability. Building a model trained on such a dataset might, therefore, be biased to a specific flavour of sarcasm, being unable to capture other flavours, increasing the risk of false negatives and limiting the ability of trained models to generalise.
Finally, if a text does not contain the predefined tags, it is considered non-sarcastic. This is a strong and problematic assumption that can lead to false negatives.
%
\begin{table}[t]
\centering
\small
    \begin{tabular}{@{}lcc}
        \toprule
         & with tag & without tag\\\hline
        annot. sarcastic & \cellcolor{green!45!gray!45} 345 & \cellcolor{red!45!gray!45} 26 \\
        annot. non-sarcastic & \cellcolor{red!45!gray!45} 486 & \cellcolor{green!45!gray!45} 975 \\ 
        \bottomrule
    \end{tabular}
    \caption{
    The agreement between manual annotation and the presence of sarcasm tags in the Riloff dataset, as discussed in Section~\ref{section:motivation:manual-labelling}.
    }
    \label{table:riloff-disagreement}
\end{table}
\subsection{Limitations of Manual labelling}
\label{section:motivation:manual-labelling}
The main limitation of manual labelling is the absence of evidence on the intention of the author of the texts that are being labelled. Annotator perception may be different to author intention, in light of studies that point out how sarcasm perception varies across socio-cultural contexts~\citep{Context2,Context3}.

\citet{joshi-riloff} provide more insight into this problem on the Riloff dataset. They present the dataset, initially labelled by Americans, to be labelled by Indians who are trained linguists.
They find higher disagreement between Indian and American annotators, than between annotators of the same nationality. Furthermore, they find higher disagreement between pairs of Indian annotators, indicating higher uncertainty, than between pairs of American annotators. They attribute these results to socio-cultural differences between India and the United States. They conclude that sarcasm annotation expands beyond linguistic expertise and is dependent on considering such factors.

Labels provided by third-party annotators might therefore not reflect the sarcastic intention of the authors of the texts that are being labelled, making this labelling method sub-optimal for capturing intended sarcasm.
To investigate this further, we looked at the Riloff dataset, which is published as a list of labelled tweet IDs. We could only retrieve 1,832 tweets, the others being removed from Twitter. We looked at the agreement between the presence of tags and manual annotation. Table~\ref{table:riloff-disagreement} shows the results. We notice that 58\% of the tweets that contained the predefined hashtags were labeled non-sarcastic. 
This disagreement between distant supervision and manual annotation provides further evidence to doubt the ability of the latter to capture intended sarcasm, at least not the flavor that distant supervision might capture. We could not perform the same analysis on the SemEval-2018 dataset because only the text of the tweets is provided, hashtags are filtered out, and tweet IDs are not available.

As we have shown, both labelling methods use a proxy for labelling sarcasm, in the form of predefined tags, predefined sources, or third-party annotators. As such, they are unable to capture the sarcastic intention  of the authors of the texts they label, generating both false positives and false negatives.
Our objective is to create a noise-free dataset of texts labelled for sarcasm, where labels reflect the sarcastic intention of the authors.
%
%
\section{Data Collection}
\label{section:data-collection}
%
\subsection{Collecting Sarcastic Tweets}
\label{section:collection:intended-sarcasm}
We designed an online survey where we asked Twitter users to provide links to one sarcastic and three non-sarcastic tweets that they had posted in the past, on their timeline, or as replies to other tweets. We made it clear that the tweets had to be their own and no retweets were allowed. We further required that the tweets should not include references to multimedia content or, if such content was referred, it should not be informative in judging sarcasm.

For each sarcastic tweet, users had to
provide, in full English sentences, an \emph{explanation} of why it was sarcastic and a \emph{rephrase} that would convey the same message non-sarcastically. This way, we aimed to prevent them from misjudging the sarcastic nature of their previous tweets under experimental bias.
Finally, we asked for their age, gender, birth country and region, and current country and region. We use the term \emph{response} to refer to all data collected from one submission of the survey.

To ensure genuine responses, we implemented the following quality control steps:
\begin{itemize}[noitemsep]
    \item The provided links should point to tweets posted no sooner than 48 hours before the submission, to prevent users from posting and providing tweets on the spot;
    \item All tweets in a response should come from the same account;
    \item Tweets cannot be from verified accounts or accounts with more than 30K followers to avoid getting tweets from popular accounts and claiming to be personal tweets~\footnote{The initial number was set to 5K, but some workers asked us to raise it since they had more followers.}.
    \item Tweets should contain at least 5 words, excluding any hashtags and URLs; 
    \item Links to tweets should not have been submitted in a previous response;
    \item Responses submitted in less than three minutes are discarded.
\end{itemize}
Each contributor agreed on a consent form before entering the survey, which informed them that only the IDs of the tweets they provide will be made public, to allow them to delete a tweet anytime and thus be in control of their own privacy in the future. They have agreed that we may collect public information from their profile, which is accessible via the Twitter API as long as the tweets pointed to by the provided IDs are not removed.

We published our survey on multiple crowdsourcing platforms, including Figure-Eight (F8), Amazon Mechanical Turk (AMT) and Prolific Academic (PA)\footnote{AMT: \url{www.mturk.com}, PA: \url{prolific.ac}, F8: \url{www.figure-eight.com}}. 
We could not get any quality responses from F8. In fact, most of our quality control steps were developed over multiple iterations on F8. On AMT, we retrieved some high quality responses, but, unfortunately, AMT stopped our job, considering that getting links to personal tweets of participants violates their policy.
We collected the majority of responses on PA.
\subsection{Labelling Sarcasm Categories}
\label{section:collection:sarcasm-categories}

We then asked a trained linguist to inspect each collected sarcastic tweet, along with the explanation provided by the author and the non-sarcastic rephrase, in order to validate the quality of the response and further assign the tweet to one of the following categories of \emph{ironic speech} defined by \citet{sarcasm-categories}:
\begin{enumerate}[noitemsep]
    \item\emph{sarcasm}: tweets that contradict the state of affairs and are critical towards an addressee;
    \item\emph{irony}: tweets that contradict the state of affairs but are not obviously critical towards an addressee;
    \item\emph{satire}: tweets that appear to support an addressee, but contain underlying disagreement and mocking;
    \item\emph{understatement}: tweets that undermine the importance of the state of affairs they refer to;
    \item\emph{overstatement}: tweets that describe the state of affairs in obviously exaggerated terms;
    \item\emph{rhetorical question}: tweets that include a question whose invited inference (implicature) is obviously contradicting the state of affairs;
    \item\emph{invalid}: tweets for which the explanation provided by their authors is unclear/unjustified. These were excluded from the dataset.
\end{enumerate}
%
\subsection{Collecting Third-Party Labels}
\label{section:collection:perceived-sarcasm}
In this part, we decided to replicate the manual annotation approach presented in previous research~\cite{Riloff,abercrombie2,van-hee-etal-2018-semeval} on part of our dataset, which we consider later as the test set, and compare the resulting \emph{perceived sarcasm} labels to the \emph{intended sarcasm} labels collected from the authors of the tweets.
Our aim was to estimate the human performance in detecting sarcasm as intended by the authors.

When collecting perceived sarcasm labels, we aimed to reduce noise caused by variations in how sarcasm is defined across socio-cultural backgrounds. Previous studies have shown gender~\citep{Context3} and country~\citep{joshi-riloff} to be the variables that are most influential on this definition. Based on their work, we made sure all annotators shared the same values for these variables.
We used PA to collect three annotations for each tweet in the iSarcasm dataset, and considered the dominant one as the label, which follows the same procedure as with building the Riloff dataset~\cite{Riloff}.
%
%
\section{Data Statistics and Analysis}
\label{section:data-statistics-and-analysis}
\begin{table*}[t]
    \centering
    \begin{tabular}{@{}cc|cccccc@{}}
        \toprule
            \multicolumn{2}{c|}{overall} & \multicolumn{6}{c}{sarcasm category}\\\hline
            sarcastic & non-sarcastic & sarcasm & irony & satire & underst. & overst. & rhet. question\\
            777 & 3,707 & 324 & 245 & 82 & 12 & 64 & 50 \\
        \bottomrule
    \end{tabular}
    \caption{Distribution of sarcastic tweets into the categories introduced in Section~\ref{section:collection:sarcasm-categories}.}
    \label{table:categories}
\end{table*}
\begin{table*}[h]
    \centering
    \footnotesize
    \begin{tabular}{@{}p{1.1cm}p{4.5cm}p{4.5cm}p{4cm}c@{}}
        \toprule
            \textbf{category} & \textbf{tweet text} & \textbf{explanation} & \textbf{rephrased}\\\midrule
            sarcasm & Thank @user for being so entertaining at the Edinburgh signings! You did not disappoint! I made my flight so will have plenty time to read @user & I went to a book signing and the author berated me for saying I was lying about heading to Singapore straight after the signing & I would have said 'here is the proof of my travel, I am mad you embarassed me in front of a large audience'!\\\midrule
            %
            irony & Staring at the contents of your fridge but never deciding what to eat is a cool way to diet & I wasn't actually talking about a real diet. I was making fun of how you never eat anything just staring at the contents of your fridge full of indecision. & I'm always staring at the contents of my fridge and then walking away with nothing cause I can never decide.\\\midrule
            satire & @mizzieashitey @PCDPhotography Totally didn’t happen, it’s a big conspiracy, video can be faked....after all, they’ve been faking the moon landings for years & It's an obvious subversion of known facts about mankind's space exploration to date that are nonetheless disputed by conspiracy theorists. & It's not a conspiracy, the video is real... after all, we've known for years that the moon landings happened.\\\midrule
            underst. & @user @user @user Still made 5 grand will do him for a while & The person I was tweeting to cashed out 5k in a sports accumulator - however he would've won 295k. "Still made 5k will do him for a while" is used to underplay the devastation of losing out. & He made 5 grand, but that will only last him a month.\\\midrule
            overst. & the worst part about quitting cigarettes is running into people you went to high school with at a vape shop & There are many things that are actually harder about quitting cigarettes than running into old classmates. & Running into old classmates at a vape shop is one of the easier things you have to deal with when you quit cigarettes.\\\midrule
            rhetorical question & @user do all your driver's take a course on how to \#tailgate! & Drivers don't have to take a course on how to tailgate its just bad driving on their part. & Could you ask your drivers not to tailgate other people on the roads please?
            \\
        \bottomrule
    \end{tabular}
    \caption{Examples of sarcastic tweets from our datasets along with the explanations that authors gave to what makes their tweets sarcastic (explanation) and how they can rephrase them to be non-sarcastic (rephrased).}
    \label{table:examples}
\end{table*}
\subsection{iSarcasm Dataset}
\begin{figure}[t]
    \centering
    \includegraphics[width=0.9\columnwidth]{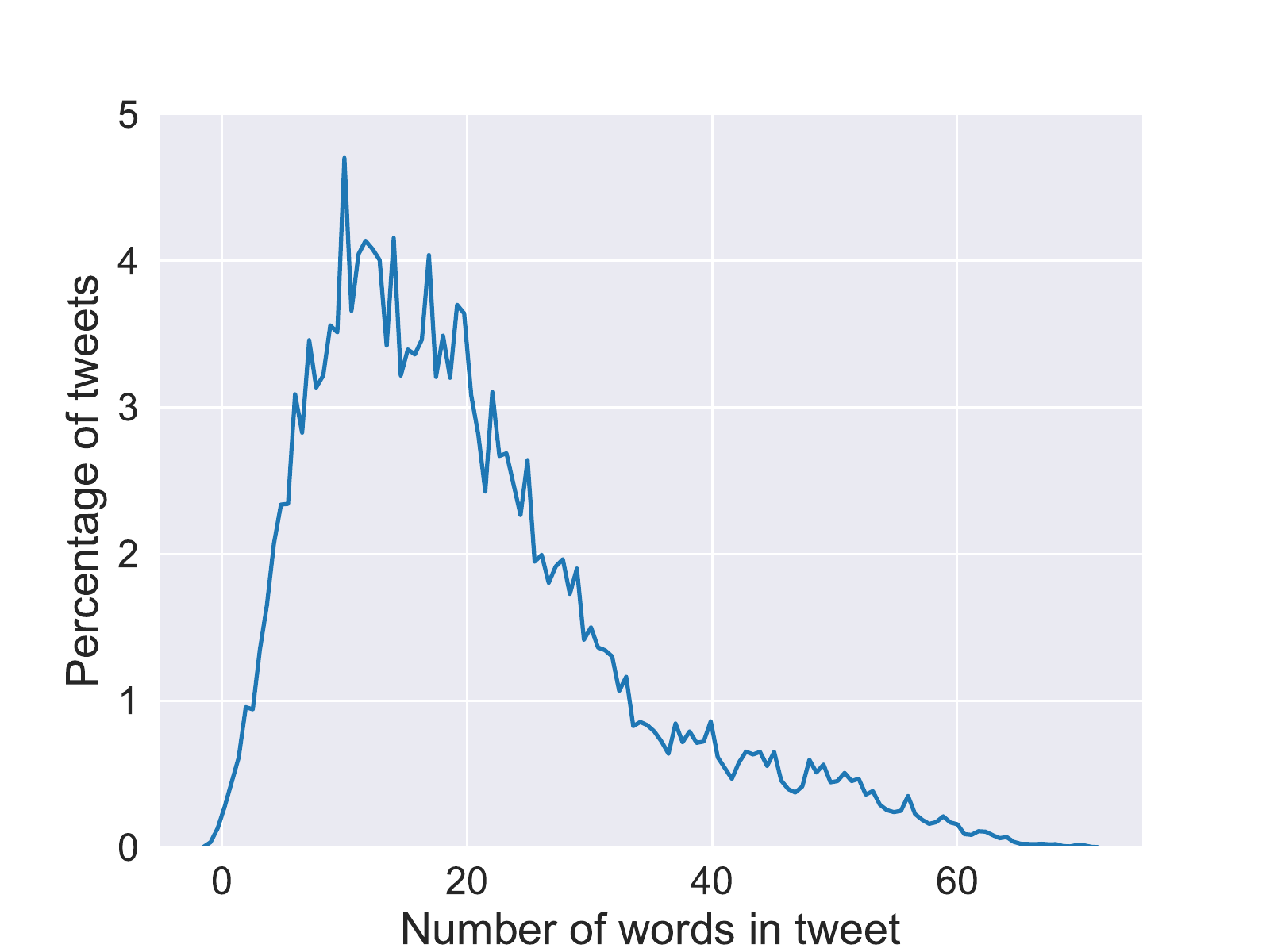}
    \caption{Tweet length distribution across iSarcasm.}
    \label{fig:tweet-len-dist}
\end{figure}

We received 1,236 responses to our survey. Each response contained four tweets labelled for sarcasm by their author, one sarcastic and three non-sarcastic. As such, we received 1,236 sarcastic and 3,708 non-sarcastic tweets.
We filtered tweets using the quality control steps described in Section~\ref{section:data-collection}, and further disregarded all tweets that fall under the \emph{invalid} category. The resulting dataset is what we call iSarcasm, containing 777 sarcastic and 3,707 non-sarcastic tweets. For each sarcastic tweet, we have its author's explanation as to why it is sarcastic, as well as how they would rephrase the tweet to be non-sarcastic. The average length of a tweet is around 20 words. \minorchange{Figure~\ref{fig:tweet-len-dist} shows the tweet length distribution across iSarcasm.} The average length of explanations 21 words, and of rephrases 14 words. Over 46\% of the tweets were posted in 2019, over 83\% starting with 2017, and the earliest in 2008.

\begin{figure}[t]
    \includegraphics[width=0.9\columnwidth]{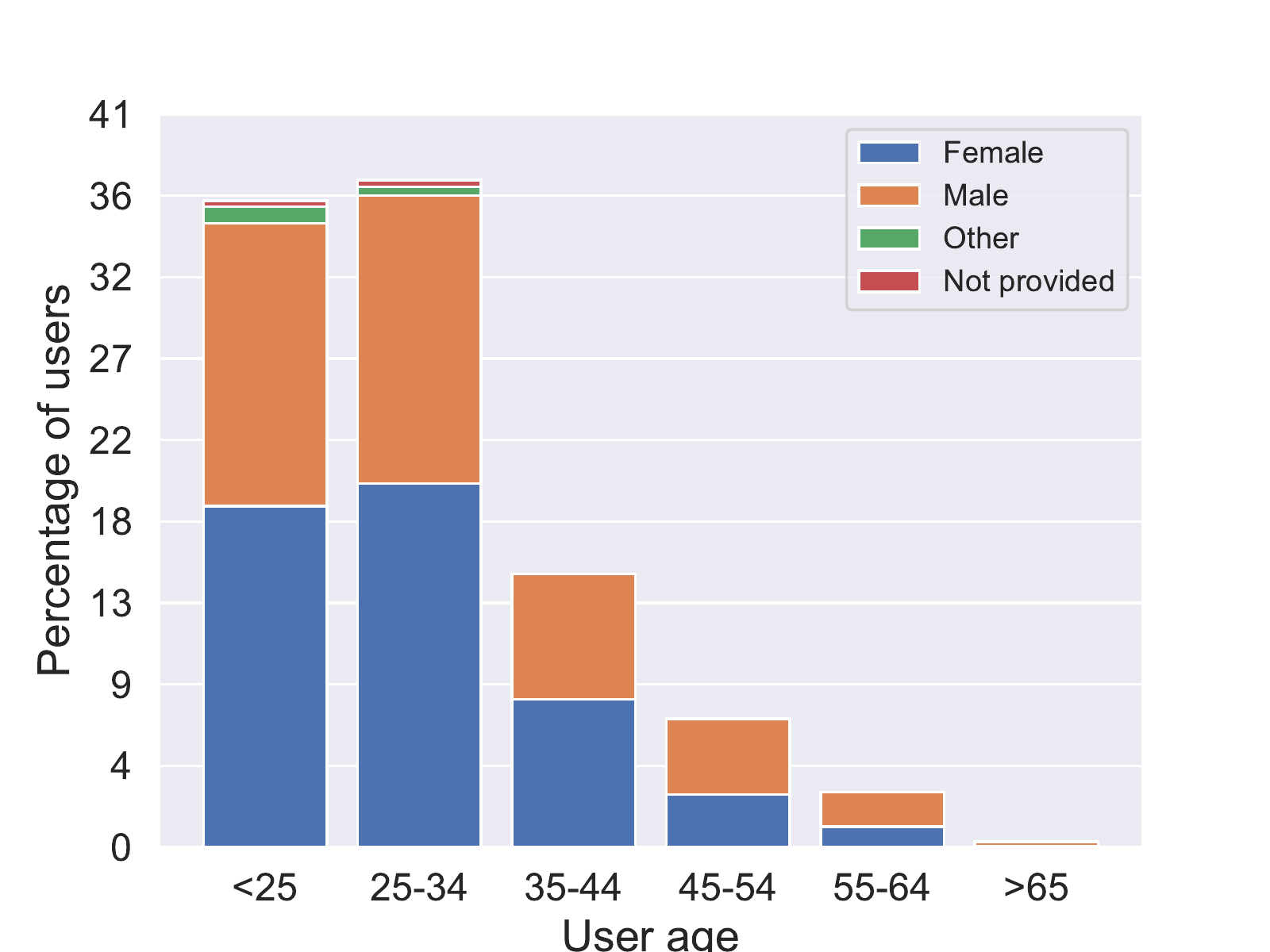}
    \caption{Age and gender distributions across the Twitter users who provided tweets in iSarcasm.}
    \label{fig:user-age-gender-dist}
\end{figure}

Among the contributors who filled our survey and provided the tweets, 56\% are from the UK and 41\% from the US, while 3\% are from other countries such as Canada and Australia. 51\% are females, and over 72\% are less than 35 years old. \minorchange{Figure~\ref{fig:user-age-gender-dist} shows the age and gender distributions across contributors.}

In iSarcasm, we investigated the presence of the hashtags \#sarcasm, \#sarcastic, and others often used to mark sarcasm in previous distant supervision datasets. None of our tweets contains any of those tags, which confirms one of our discussed limitations of this approach, that the lack of tags should not be associated with lack of sarcasm, and that these tags might capture only one flavor of sarcasm, not sarcasm present on social media in general.

Regarding the categories of sarcasm, assigned by the linguist to the sarcastic tweets, Table~\ref{table:categories} shows the distribution of the tweets into these categories. As shown, sarcasm and irony are the largest two categories (73\%), while understatement is the smallest one (with only 12 tweets).
Table~\ref{table:examples} shows examples of the sarcastic tweets, along with the explanations and rephrases provided by the authors.

iSarcasm is published as two files, a training set and a test set, containing 80\% and 20\% of the examples chosen at random, respectively. Each file contains tweet IDs along with corresponding intended sarcasm labels. For sarcastic tweets we also provide the category of ironic speech they belong to. This is in accordance with the consent form that the contributors have agreed to, whose privacy we take seriously.
Nonetheless, we still offer the tweets text along with the explanations and rephrases of the sarcastic tweets provided by the authors for free for research purposes, under an agreement that protects the privacy of our contributors. 
\subsection{Third-Party Labels}
\label{section:statistics:3rd-party}
As we mentioned earlier, we collected three third-party labels for each tweet in the test set of iSarcasm. Using Cohen's kappa ($\kappa$; \citet{cohen}) as a measure, the pairwise inter-annotator agreement (IAA) scores were $\kappa_{12}=0.37$, $\kappa_{13}=0.39$ and $\kappa_{23}=0.36$, which highlights the high subjectivity of the task.
We used majority voting to select the final perceived sarcasm label for each tweet. Table~\ref{table:ip-disagreement} shows the disagreement between the intended and perceived labels.
As shown, 30\% of the sarcastic tweets were unrecognised by the annotators, while 45\% of the tweets perceived as sarcastic were actually not intended to be sarcastic by their authors. This supports our argument that third-party annotation for sarcasm should not be trusted.
%
%
\begin{table}
\centering
    \begin{tabular}{@{}lcc}
        \toprule
         & perc. sarc.  & perc. non-sarc.\\\hline
        int. sarc.    & \cellcolor{green!45!gray!45} 61 & \cellcolor{red!45!gray!45} 26\\
        int. non-sarc.& \cellcolor{red!45!gray!45} 50 & \cellcolor{green!45!gray!45} 322 \\ 
        \bottomrule
    \end{tabular}
    \caption{
    The agreement between intended labels (\emph{int.}), provided by the authors, and perceived labels, provided by third-party annotators, (\emph{perc.}) on iSarcasm test set.}
    \label{table:ip-disagreement}
\end{table}
%
%
\section{Detecting Intended Sarcasm}
\label{section:baselines}
In the following, we examine the effectiveness of state-of-the-art sarcasm detection models on iSarcasm. We aim to investigate their ability to detect intended sarcasm rather than sarcasm labeled using distant supervision or manual annotation. As we have shown, these labelling methods could produce noisy labels.
We experiment with those models that have achieved state-of-the-art results on previous benchmark datasets for sarcasm detection.
\subsection{Baseline Datasets}
We consider four previously published datasets. Two of them, Riloff~\citep{Riloff} and SemEval-2018~\citep{van-hee-etal-2018-semeval}, were labeled via a hybrid approach of distant supervision for initial collection and manual annotation for actual labelling. The other two datasets, Ptacek~\citep{ptacek} and SARC~\citep{khodak-2017}, are labeled using distant supervision.
As mentioned earlier, we managed to collect 1,832 tweets from the Riloff dataset.
SemEval-2018 is a balanced dataset consisting of 4,792 tweets. 
For the Ptacket dataset, we collected 27,177 tweets out of the 50K published tweet IDs.
Finally, The SARC datasets consists of Reddit comments. In a setting similar to~\citet{cascade} who publish state-of-the-art results on this dataset, we consider two variants of SARC. SARC-balanced contains 154,702 comments with the same number of sarcastic and non-sarcastic comments, while SARC-imbalanced contains 103,135‬ comments with a ratio of about 20:80 between sarcastic and non-sarcastic comments.
\subsection{Sarcasm Detection Models}
\label{section:baselines:models}
\paragraph{Riloff and Ptacek datasets} We replicate the models implemented in~\cite{tay-att}, who report state-of-the-art results on Riloff and Ptacek. These models are:
\textbf{LSTM} first encodes the tweet with a recurrent neural network with long-term short memory units (LSTM; \citet{lstm}), then adds a binary softmax layer to output a probability distribution over labels (sarcastic or non-sarcastic) and assigns the most probable label. \minorchange{It has one hidden layer of dimension 100.}
\textbf{Att-LSTM} adds an attention mechanism on top of the LSTM, in the setting specified by~\citet{yang-etal-2016-hierarchical}. \minorchange{In particular, it uses the attention mechanism introduced by \citet{bahdanau-attention} of dimension 100.}
\textbf{CNN} encodes the tweet with a convolutional neural network (CNN) \minorchange{with 100 filters of size 3} and provides the result to feed-forward network with a final binary softmax layer, choosing the most probable label.
\textbf{SIARN} (Single-Dimension Intra-Attention Network;~\citet{tay-att}) is the model that yields the best published performance on the Riloff dataset. It relies on the assumption that sarcasm is caused by linguistic incongruity between words. It uses an intra-attention mechanism~\citep{shen2018disan} between each pair or words to detect this incongruity.
\textbf{MIARN} (Multi-Dimension Intra-Attention Network;~\citet{tay-att}) reports the best results on the Ptacek dataset. In addition to SIARN, MIARN allows multiple intra-attention scores for each pair of words to account for multiple possible meanings of a word when detecting incongruity. We use an implementation of MIARN similar to that described by its authors. \minorchange{We set the dimension of all hidden layers of \textbf{SIARN} and \textbf{MIARN} to 100.}
\paragraph{SARC datasets}\citet{cascade} report the best results on SARC-balanced and SARC-imbalanced, to our knowledge. However, they model both the content of the comments as well as contextual information available about the authors. In this paper we only focus on content modelling, using a convolutional network (\textbf{3CNN}) in a setting similar to what they describe. \minorchange{\textbf{3CNN} uses three filter types of sizes 3, 4, and 5, with 100 filters for each size.}
\paragraph{SemEval-2018 dataset} The SemEval dataset contains two types of labels for each tweet: binary labels that specify whether the tweet is sarcastic or not; and labels with four possible values, specifying the type of sarcasm present\footnote{We use ``sarcasm'' to mean what they refer to as ``verbal irony''.}. \citet{semeval-best} report the best results on both tasks with their \textbf{Dense-LSTM} model. Given a tweet, the model uses a sequence of four LSTM layers to compute a hidden vector $H$. $H$ is then concatenated with a tweet embedding $S$ computed in advance by averaging embeddings of all words inside using the pre-trained embeddings provided by \citet{bravo-embeddings}. $H$ and $S$ are further concatenated with a sentiment feature vector of the tweet computed in advance using the \emph{weka} toolkit~\citep{mohammad-bravo-marquez-2017-wassa}, by applying the \emph{TweetToLexiconFeatureVector}~\citep{BRAVOMARQUEZ201486} and \emph{TweetToSentiStrengthFeatureVector}~\citep{sentistrength} filters.
\minorchange{The authors of Dense-LSTM train the network in a multitask setting on the SemEval dataset~\citep{van-hee-etal-2018-semeval} to predict three components: the binary sarcasm label, one of the four types of sarcasm, and the corresponding hashtag, if any, that was initially used to mark the tweet as sarcastic, out of \#sarcasm, \#sarcastic, \#irony and \#not. \citet{semeval-best} report an F-score of 0.674 using a fixed dropout rate of 0.3 in all layers. They further report an F-score of 0.705 by averaging the performance of 10 Dense-LSTM models, varying the dropout rate to random values between 0.2 and 0.4. We implement and train it to only predict the binary sarcasm label, to make it applicable to iSarcasm and make the results on SemEval-2018 and iSarcasm comparable.}

For each previous dataset, we implemented the models reported previously to achieve the best performance on that dataset, and made sure our implementations achieve similar performance to the published one. This is confirmed in Table~\ref{table:results-of-soa-models}, providing confidence in the correctness of our implementations.
%
%
\begin{table}
    \centering
    \small
    \begin{tabular}{@{}llcc@{}}
        \toprule
        \textbf{Dataset} & \textbf{Model} & 
        \textbf{published} & \textbf{our impl.}\\\midrule
        Riloff  & LSTM & 0.673 & 0.669 \\
                & Att-LSTM & 0.687 & 0.679\\
                & CNN & 0.686 & 0.681 \\
                & SIARN & 0.732 & 0.741 \\
                & MIARN & 0.701 & 0.712 \\\midrule
        Ptacek  & LSTM & 0.837 & 0.837 \\
                & Att-LSTM & 0.837 & 0.841 \\
                & CNN & 0.804 & 0.810 \\
                & SIARN & 0.846 & 0.864 \\
                & MIARN & 0.860 & 0.874 \\\midrule
        SARC-balanced & 3CNN & 0.660 & 0.675 \\
        SARC-unbalanced & 3CNN & 0.780 & 0.788 \\\midrule
        SemEval-2018 & Dense-LSTM & 0.674 & 0.666\\
        \bottomrule
    \end{tabular}
    \caption{F-score yielded by our implementations of state-of-the-art models on previous datasets, compared to published results on those datasets.}
    \label{table:results-of-soa-models}
\end{table}
\renewcommand{\arraystretch}{1.1}
\begin{table}[t!]
    \centering
    \begin{tabular}{@{}lccc@{}}
        \toprule
        \textbf{Model} & Precision & Recall & F-score \\
        \midrule
        Manual Labelling & 0.550 & 0.701 & \textbf{0.616}\\
        \hline
        LSTM  & 0.217 & 0.747 & 0.336 \\
        Att-LSTM & 0.260 & 0.436 & 0.325 \\
        CNN   & 0.261 & 0.563 & 0.356 \\
        SIARN & 0.219 & 0.782 & 0.342 \\
        MIARN & 0.236 & 0.793 & \textbf{0.364} \\
        3CNN  & 0.250 & 0.333 & 0.286 \\
        Dense-LSTM & 0.375 & 0.276 & 0.318 \\
        \bottomrule
    \end{tabular}
    \caption{Experimental results on iSarcasm. \emph{Manual Labelling} shows the results using the perceived sarcasm labels provided by third-party human annotators.}
    \label{table:results}
\end{table}
%
%
\subsection{Results and Analysis}
\label{section:baselines:results}
Table~\ref{table:results} reports precision, recall and f-score results on the test set of iSarcasm using the detection models discussed, alongside third-party annotator performance.
As shown, all the models perform significantly worse than humans, who achieve an F-score of only 0.616. 
MIARN is the best performing model with a considerably low F-score of 0.364, compared to its performance on the Riloff and Ptacek datasets (0.741 and 0.874 F-scores respectively). 3CNN achieves the lowest performance on iSarcasm with an F-Score of 0.286 compared to 0.675 and 0.788 on SARC balanced and imbalanced, respectively. Similarly,
Dense-LSTM achieves 0.318, compared to 0.666 on SemEval-2018.

Previous models that achieved high performance in detecting sarcasm on datasets sampling perceived sarcasm (third-party labels) or hash-tagged sarcasm (distant supervision) have failed dramatically to detect sarcasm as meant by its author.
This motivates the need to develop more effective methods for detecting intended sarcasm. Potentially, building models that account for sociocultural traits of the authors (available on, or inferred from, their Twitter profiles), or consider other contextual elements to judge the sarcasm in our dataset~\citep{Context2}. Previous research has considered certain contextual elements~\citep{bamman-2015,amir,cascade,oprea-2019}, but only on sarcasm captured by previous labelling methods.

We believe the iSarcasm dataset, with its novel method of sampling sarcasm as intended by its authors, shall revolutionise research in sarcasm detection in the future; and open the direction for new sub-tasks, such as sarcasm category prediction, and sarcasm decoding/encoding, using information found both in the tweets themselves, and in the explanations and rephrases provided by the authors, available with each sarcastic tweet in the dataset.
%
%
%
%
\section{Conclusion and Future Work}
\label{section:conclusion}
In this paper, we presented iSarcasm, a dataset of intended sarcasm consisting of 4,484 tweets labeled and explained by their authors, and further revised and categorised by an expert linguistic. We believe this dataset will allow future work in sarcasm detection to progress in a setting free of the noise found in existing datasets. We saw that computational models perform poorly in detecting sarcasm in the new dataset, indicating that the sarcasm detection task might be more challenging compared to how it was seen in earlier research. We aim to promote research in sarcasm detection, and to encourage future investigations into sarcasm in general and how it is perceived across cultures.

\section*{Acknowledgments}
This work was supported in part by the EPSRC
Centre for Doctoral Training in Data Science,
funded by the UK Engineering and Physical Sciences Research Council (grant EP/L016427/1);
the University of Edinburgh; and The Financial
Times.
\bibliography{index}
\bibliographystyle{acl_natbib}
\end{document}


\maketitle
%
%
\section{Model Implementation Details}
In Section 6 in the paper, we tested state-of-the-art models on our dataset, showing poor performance. To make sure this is not due to bugs on our side, we made sure our implementations of those models achieved similar results to published ones on the respective datasets. We show specific comparisons in Table 6 in the paper.

We now provide implementation details for the models used. \textbf{LSTM} has one hidden layer and dimension 100. On top of this, \textbf{Att-LSTM} uses the attention mechanism introduced by \citet{bahdanau-attention} of dimension 100. \textbf{CNN} has a convolutional layer with 100 filters of size 3. We set the dimension of all hidden layers of \textbf{SIARN} and \textbf{MIARN} to 100. \textbf{3CNN} uses three filters of sizes 3, 4, and 5, with 100 filters for each size.
For \textbf{Dense-LSTM}, the original authors~\citep{semeval-best} train it in a multitask setting on the SemEval dataset~\citep{van-hee-etal-2018-semeval} to predict three components: the binary sarcasm label, one of the four types of sarcasm, and the corresponding hashtag, if any, that was initially used to mark the tweet as sarcastic, out of \#sarcasm, \#sarcastic, \#irony and \#not. \citet{semeval-best} report an F-score of 0.674 using a fixed dropout rate of 0.3 in all layers. They further report an F-score of 0.705 by averaging the performance of 10 Dense-LSTM models, varying the dropout rate to random values between 0.2 and 0.4. We implement and train it to only predict the binary sarcasm label, to make it applicable to iSarcasm and make the results on SemEval and iSarcasm comparable. We use a fixed dropout rate of 0.3.

In preprocessing, we use the spaCy library\footnote{https://spacy.io/api/tokenizer} for tweet tokenization. We then replace all tokens that only appear once in the corpus with $\langle$unk$\rangle$. We also replace all handles with a $\langle$user$\rangle$, ellipses with the $\langle$ellipsis$\rangle$ token, and numbers with the $\langle$number$\rangle$ token. We further remove all punctuation except ``.'', ``!'', ``?'', ``('',  and ``)''. We represent each tweet as a sequence of word vectors initialized using GloVe embeddings~\citep{glove2014} of dimension 100.

We group our data into batches of 128 examples, train all models for 30 epochs using the Adam optimizer~\citep{adam} with an initial learning rate of $10^{-3}$.
%
\section{iSarcasm Statistics}
\begin{figure}[t]
    \begin{subfigure}{.45\textwidth}
    \centering
    \includegraphics[width=\linewidth]{img/tweet-len-dist.pdf}
    \caption{}
    \label{fig:tweet-len-dist}
    \end{subfigure}
    \begin{subfigure}{.45\textwidth}
    \centering
    \includegraphics[width=\linewidth]{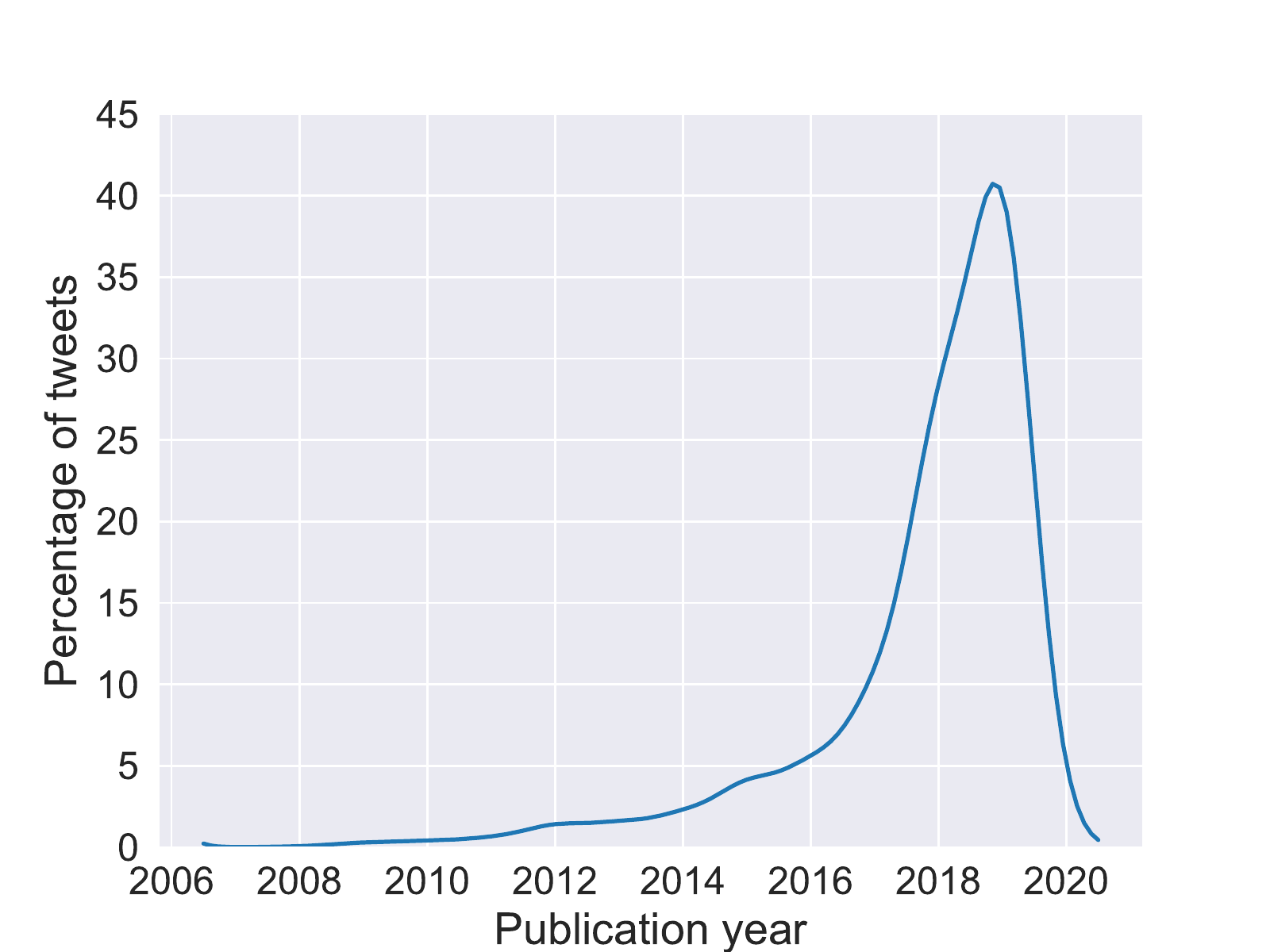}
    \caption{}
    \label{fig:tweet-timestamp-dist}
    \end{subfigure}
    \caption{Tweet length (above) and publication year (below) distribution across iSarcasm.}
    \label{}
\end{figure}

As discussed in Section 5 in the paper, iSarcasm contains 777 sarcastic and 3,707 non-sarcastic tweets in its final form, after applying all quality control steps.
The average tweet length is iSarcasm is 20 words. Figure~\ref{fig:tweet-len-dist} shows the tweet length distribution across iSarcasm. Most tweets were published in 2019. Figure~\ref{fig:tweet-timestamp-dist} shows the publication year distribution across the tweets.

\begin{figure}[t]
    \begin{subfigure}{.45\textwidth}
    \centering
    \includegraphics[width=\linewidth]{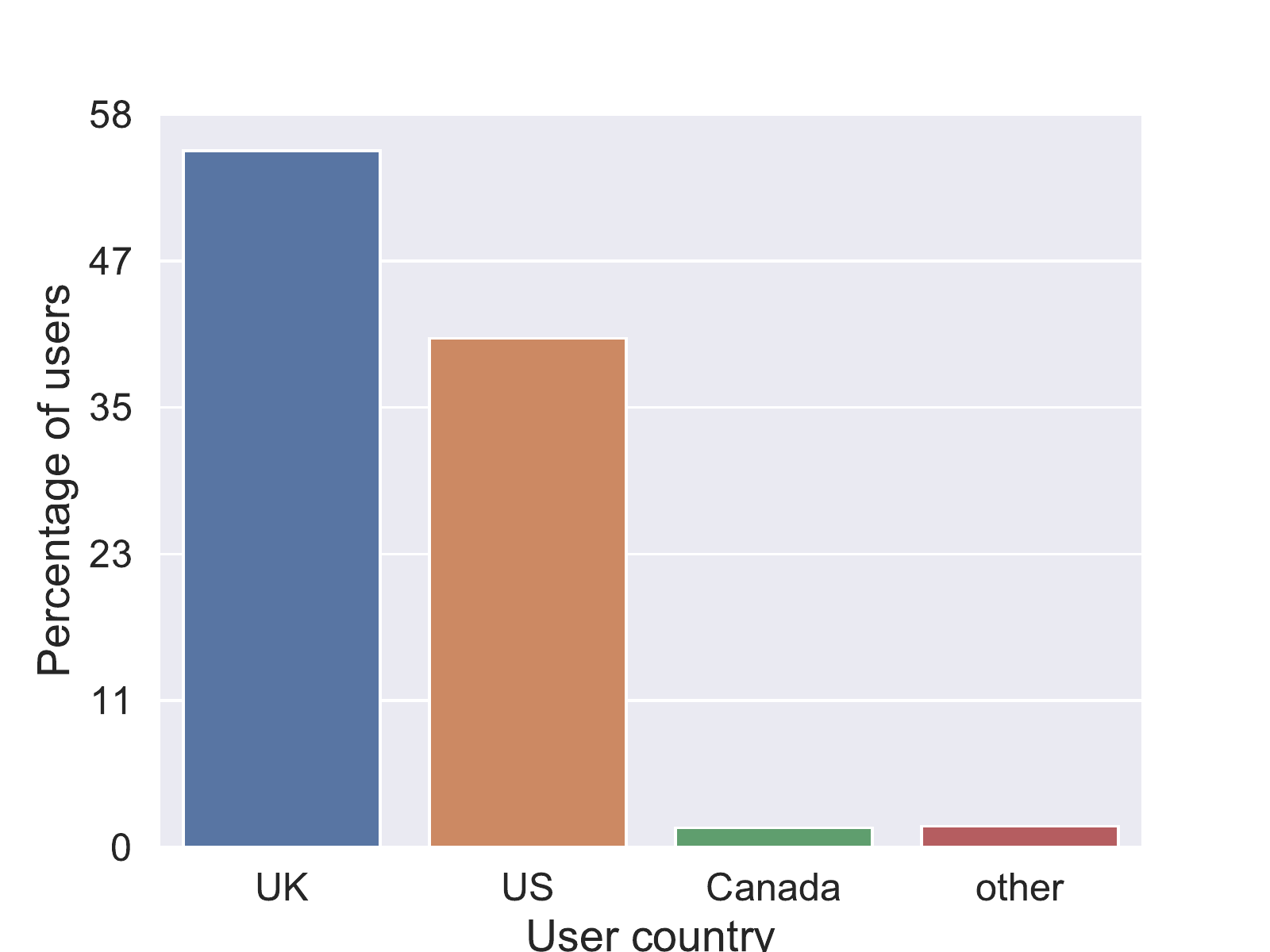}
    \caption{}
    \label{fig:country-dist}
    \end{subfigure}
    \begin{subfigure}{.45\textwidth}
    \centering
    \includegraphics[width=\linewidth]{img/age-gender-dist.pdf}
    \caption{}
    \label{fig:age-gender-dist}
    \end{subfigure}
    \caption{Country (above), and age and gender (below) distributions across the Twitter users who provided tweets in iSarcasm.}
    \label{}
\end{figure}

About 97\% of the contributors were from the UK and the US (56\% from the UK and 41\% from the US), and the others from countries such as Canada and Australia. Figure~\ref{fig:country-dist} shows the country distribution across contributors.

About 51\% of the contributors were females, and over 72\% were less than 35 years old. Figure \ref{fig:age-gender-dist} shows the age and gender distributions across contributors.
%
\section{iSarcasm Distribution}
iSarcasm is made available publicly for free for research usage. We provide two levels of details of our dataset. The two versions of iSarcasm are:

\begin{enumerate}
    \item The first version is public and  available for direct download at \url{https://bit.ly/iSarcasm}. This version does not require any license for use and contains the list of tweet IDs of the iSarcasm dataset along with the labels. It is provided in two files, each containing 80\% and 20\% of the tweets, representing the training and test sets of iSarcasm, respectively. Each line in a file has the format \verb|tweet_id|, \verb|sarcasm_label|, \verb|sarcasm_type|. \verb|sarcasm_type| is the category of sarcasm assigned by the linguist and is only defined for sarcastic tweets.
    
    \item The second version includes the additional information in iSarcasm, including the tweet text of the tweets and, for each sarcastic tweet: (a) the explanation provided by the author of why the tweet is sarcastic, and (b) the rephrase that conveys the same meaning non-sarcastically. This version is not publicly available. However, we will provide it for free for research purposes upon request, under an agreement that protects the privacy of the tweet authors, since it contains this additional information that is not public.
\end{enumerate}
%
\section{Data Collection Survey}
\label{section:data-collection-survey}
\begin{figure*}[h]
    \centering
    \includegraphics[width=\textwidth]{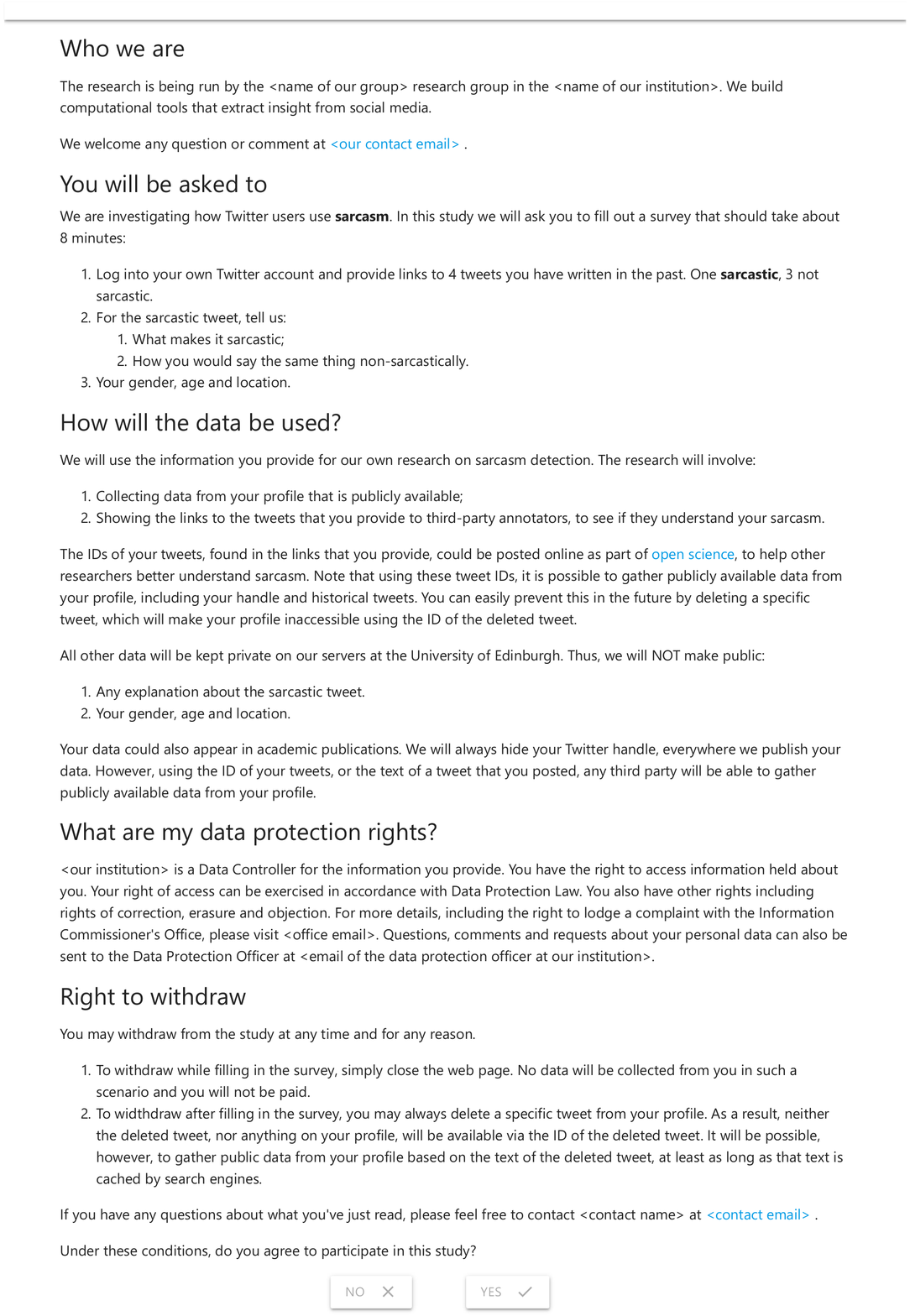}
    \caption{Consent form shown to workers on Prolific Academic, as discussed in Section~\ref{section:data-collection-survey}.}
    \label{fig:consent-form}
\end{figure*}

\begin{figure*}[h]
    \centering
    \includegraphics[width=\textwidth]{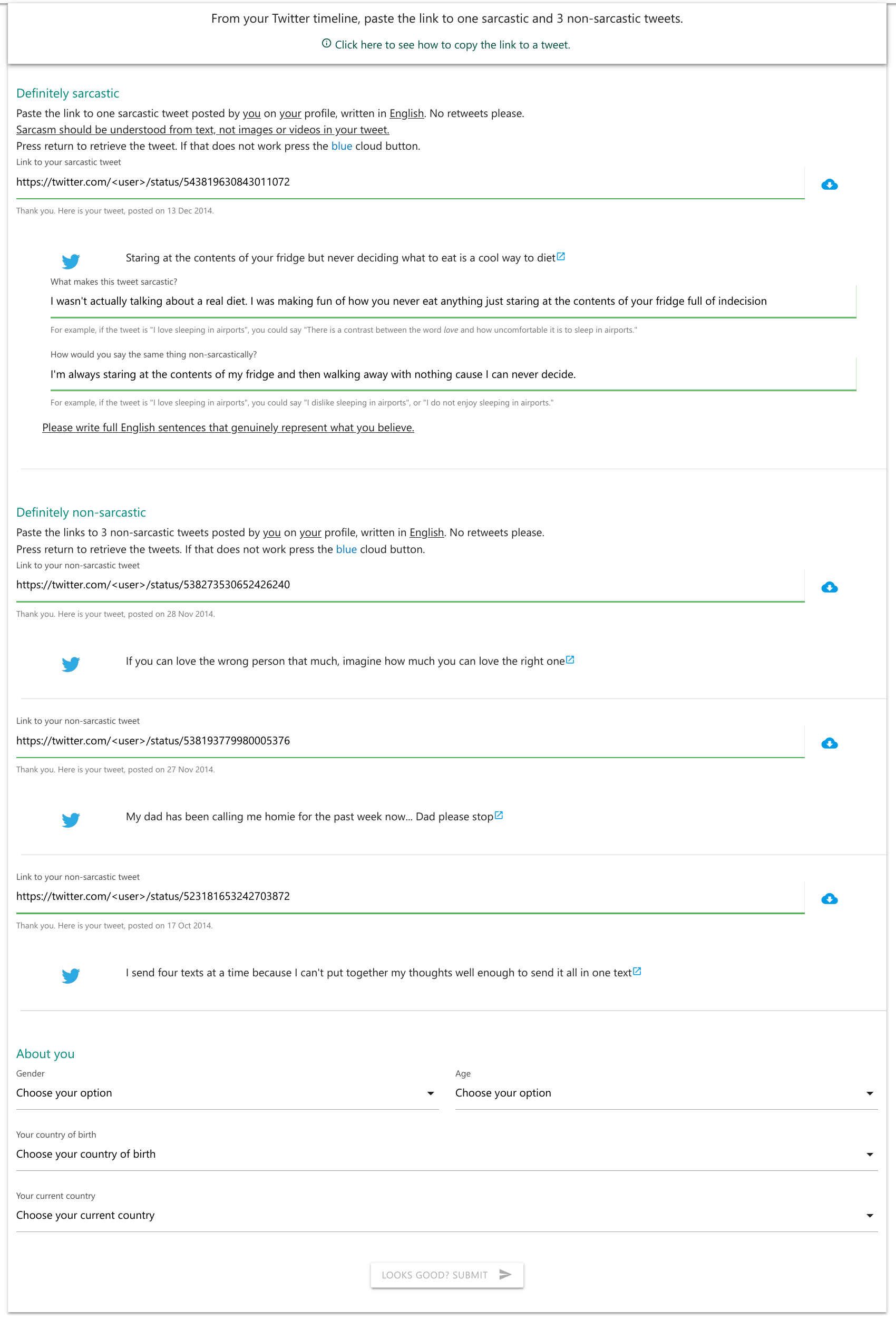}
    \caption{Data collection survey shown to workers on Prolific Academic, as discussed in Section~\ref{section:data-collection-survey}.}
    \label{fig:survey}
\end{figure*}

Figure~\ref{fig:consent-form} shows the consent form that the workers on Prolific Academic had to agree to in order to enter our survey. Figure~\ref{fig:survey} show a snapshot of the survey itself.
\bibliography{index}
\bibliographystyle{acl_natbib}